\pdfoutput=1
\documentclass[10pt,journal,cspaper,twocolumn,compsoc]{IEEEtran}

\usepackage{graphicx}
\usepackage{url}
\usepackage{hyperref}
\usepackage{amsmath}

\usepackage{mathrsfs}
\usepackage{amsmath}
\usepackage{amssymb}
\usepackage{amssymb}
\usepackage{booktabs}
\usepackage{algorithm,algorithmic,alltt}
\usepackage{times}
\usepackage{epsfig}
\usepackage{graphicx}
\usepackage{lineno}
\usepackage{color}
\usepackage{times}
\usepackage{epsfig}
\usepackage{algorithm,algorithmic,alltt}
\usepackage[font=small]{caption}
\usepackage{tabularx}
\usepackage{multirow}

\DeclareMathOperator*{\argmin}{argmin}

\renewcommand{\tabcolsep}{.1pt}
\def\swidthone{0.99\linewidth}

\def\swidththree{0.32\linewidth}
\def\swidthfour{0.24\linewidth}
\def\swidthfive{0.19\linewidth}

\def\swidthseven{0.13\linewidth}
\def\swidtheight{0.11\linewidth}

\newcommand{\figdir}{figures}
\newcommand{\figgray}{figures/more_results/grayscale}
\newcommand{\figours}{figures/more_results/ours}
\newcommand{\figgt}{figures/more_results/groundtruth}

\begin{document}
\title{Deep Colorization
    \thanks{Preliminary version of this work was published in ICCV 2015 \cite{cheng2015deep}}
    \thanks{Zezhou Cheng is with the Department of Computer Science and Engineering, Shanghai Jiao Tong University, Shanghai 200240, China (e-mail: chengzezhou@sjtu.edu.cn).}
    \thanks{Qingxiong Yang is with the Department of Computer Science at the City University of Hong Kong, Hong Kong, China (e-mail: qiyang@cityu.edu.hk).}
    \thanks{Bin Sheng is with the same Department as Zezhou Cheng (e-mail: shengbin@sjtu.edu.cn).}
    \thanks{Matlab code, trained models and more colorization results are available at the authors' website.}
}
\author{
Zezhou Cheng,~\IEEEmembership{Student Member,~IEEE,}
Qingxiong Yang,~\IEEEmembership{Member,~IEEE,}
Bin Sheng,~\IEEEmembership{Member,~IEEE,}\\
{\small\url{http://www.cs.cityu.edu.hk/~qiyang/publications/iccv15/}}
}

\maketitle

\begin{abstract}
This paper investigates into the colorization problem which converts a grayscale image to a colorful version. This is a very difficult problem and normally requires manual adjustment to achieve artifact-free quality. For instance, it normally requires human-labelled color scribbles on the grayscale target image or a careful selection of colorful reference images (e.g., capturing the same scene in the grayscale target image). Unlike the previous methods, this paper aims at a \emph{high-quality fully-automatic} colorization method. With the assumption of a perfect patch matching technique, the use of an extremely large-scale reference database (that contains sufficient color images) is the most reliable solution to the colorization problem. However, patch matching noise will increase with respect to the size of the reference database in practice. Inspired by the recent success in deep learning techniques which provide amazing modeling of large-scale data, this paper re-formulates the colorization problem so that deep learning techniques can be directly employed. To ensure artifact-free quality, a joint bilateral filtering based post-processing step is proposed. We further develop an adaptive image clustering technique to incorporate the global image information. Numerous experiments demonstrate that our method outperforms the state-of-art algorithms both in terms of quality and speed.
\end{abstract}

\begin{keywords}
Image colorization, neural networks
\end{keywords}

\section{Introduction}
Image colorization assigns a color to each pixel of a target grayscale image. Colorization methods can be roughly divided into two categories: scribble-based colorization \cite{Huang,Levin,Luan, Qu, Yatziv} and example-based colorization \cite{charpiat, chia, gupta, irony, liu,Welsh}. The scribble-based methods typically require substantial efforts from the user to provide considerable scribbles on the target grayscale images. It is thus time-assuming to colorize a grayscale image with fine-scale structures, especially for a rookie user.

To reduce the burden on user, \cite{Welsh} proposes an example-based method which is later further improved by \cite{charpiat,irony}. The example-based method typically transfers the color information from a similar reference image to the target grayscale image. However, finding a suitable reference image becomes an obstacle for a user. \cite{chia, liu} simplify this problem by utilizing the image data on the Internet and propose filtering schemes to select suitable reference images. However, they both have additional constraints.  \cite{liu} requires identical Internet object for precise per-pixel registration between the reference images and the target grayscale image. It is thus limited to objects with a rigid shape (\textit{e.g.} landmarks). \cite{chia} requires user to provide a semantic text label and segmentation cues for the foreground object. In practice, manual segmentation cues are hard to obtain as the target grayscale image may contain multiple complex objects (\textit{e.g.} building, car, tree, elephant). These methods share the same limitation $-$ their performance highly depends on the selected reference image(s).

A fully-automatic colorization method is proposed to address this limitation. Intuitively, one reference image cannot include all possible scenarios in the target grayscale image. As a result, \cite{charpiat, chia, irony, Welsh} require similar reference image(s). A more reliable solution is locating the most similar image patch/pixel in a huge reference image database and then transferring color information from the matched patch/pixel to the target patch/pixel. However, the matching noise is too high when a large-scale database is adopted in practice.

\begin{figure*}[!ht]
    \begin{center}
        \begin{tabular}{c}
        \includegraphics[width=\swidthone,height=0.33\linewidth]{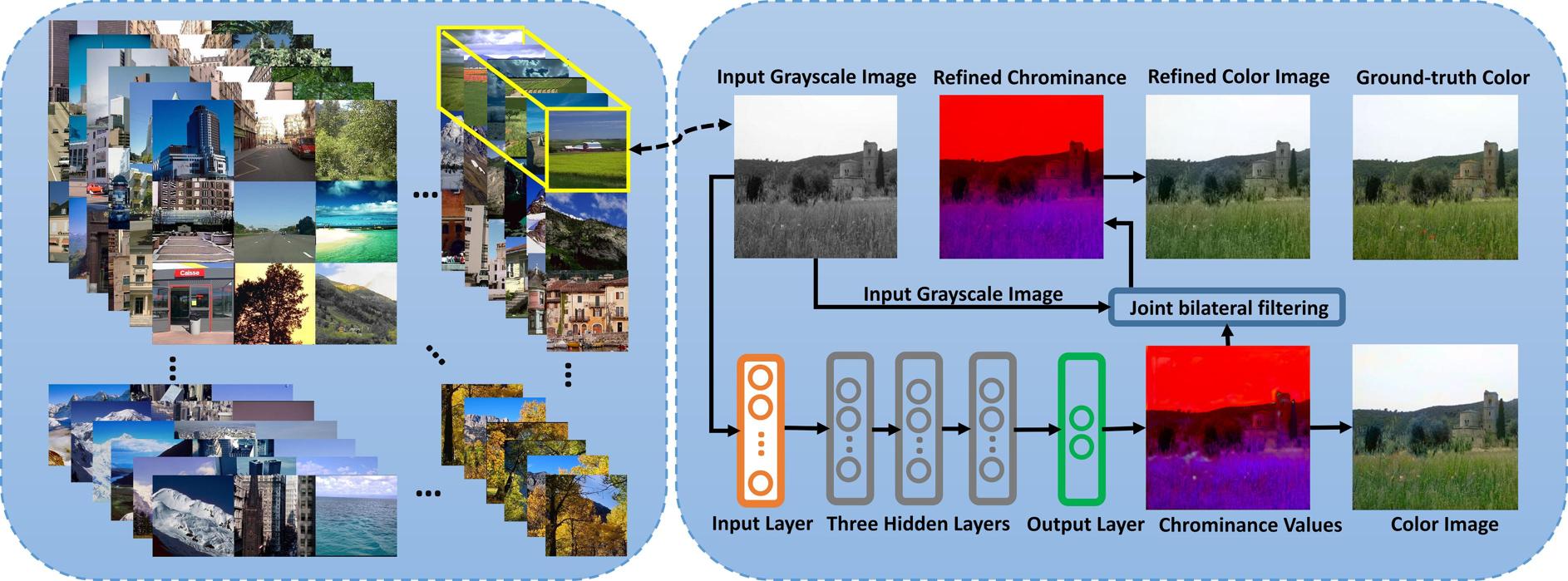}\\
        \leftline{~~~~~~~~~~~~~~(a) Reference image clusters  ~~~~~~~~~~~~~~~~~~~~~~~~~~~~~~~ (b) The proposed colorization method} \\
        \end{tabular}
    \end{center}
\caption{The adopted large reference image database and overview of the proposed colorization method. (a) shows the reference images that have been grouped into various clusters by a proposed adaptive image clustering technique. A deep neural network (DNN) will be trained for each cluster. (b) presents our colorization procedure and the architecture of the proposed DNN. Given a target grayscale, the nearest cluster and corresponding trained DNN will be explored automatically first. The feature descriptors will be extracted at each pixel and serve as the input of the neural network. Each connection between pairs of neurons is associated with a weight to be learned from a large reference image database. The output is the chrominance of the corresponding pixel which can be directly combined with the luminance (grayscale pixel value) to obtain the corresponding color value. The chrominance computed from the trained model is likely to be a bit noisy around low-texture regions. The noise can be significantly reduced with a joint bilateral filter (with the input grayscale image as the guidance).}
\label{Fig:architecture}
\end{figure*}

Deep learning techniques have achieved amazing success in modeling large-scale data recently. It has shown powerful learning ability that even outperforms human beings to some extent (\textit{e.g.} \cite{he}) and deep learning techniques have been demonstrated to be very effective on various computer vision and image processing applications including image classification \cite{krizhevsky}, pedestrian detection \cite{ouyang, zeng}, image super-resolution \cite{dong}, photo adjustment \cite{yan} etc. The success of deep learning techniques motivates us to explore its potential application in our context. This paper formulates image colorization as a regression problem and deep neural networks are used to solve the problem. A large database of reference images comprising all kinds of objects (\textit{e.g.} tree, building, sea, mountain etc.) is used for training the neural networks. Some example reference images are presented in Figure \ref{Fig:architecture} (a). Although the training is significantly slow due to the adoption of a large database, the learned model can be directly used to colorize a target grayscale image efficiently. The state-of-the-art colorization methods normally require matching between the target and reference images and thus are slow.

It has recently been demonstrated that high-level understanding of an image is of great use for low-level vision problems (\textit{e.g.} image enhancement \cite{yan}, edge detection \cite{zheng}). Because image colorization is typically semantic-aware, we propose a new semantic feature descriptor to incorporate the semantic-awareness into our colorization model.

An adaptive image clustering is proposed to incorporate the global image information to reduce the training ambiguities.

To demonstrate the effectiveness of the presented approach, we train our deep neural networks using a large set of reference images from different categories as can be seen in Figure \ref{Fig:architecture} (a). The learned model is then used to colorize various grayscale images in Figure \ref{Fig:more_results}. The colorization results shown in Figure \ref{Fig:more_results} demonstrate the robustness and effectiveness of the proposed method.

The major contributions of this paper are as follows:
\begin{enumerate}
  \item It proposes the first deep learning based image colorization method and demonstrates its effectiveness on various scenes.
  \item It carefully analyzes informative yet discriminative image feature descriptors from low to high level, which is key to the success of the proposed colorization method.
\end{enumerate}

An initial version of this work was presented in \cite{cheng2015deep}. The present work has significant differences from the earlier version. Firstly, we propose an adaptive image clustering to classify the training images according to their global information. A neural network is trained for each image cluster and the resulted neural network assemble is used to colorize the target grayscale image. Considerable qualitative and quantitative results are shown to prove that the new framework outperforms \cite{cheng2015deep} both in colorization quality and accuracy. Secondly, more analysis of the proposed model along with comparisons to the state-of-art concurrent work \cite{deshpande2015learning} is added. Thirdly, we show that the proposed model is flexible to learn various colorization styles. Additionally, we update the experimental results reported in \cite{cheng2015deep} due to changes between the preliminary and the current work.

\section{Related Work}
This section gives a brief overview of the previous colorization methods.

\textbf{Scribble-based colorization} Levin et al. \cite{Levin} propose an effective approach that requires the user to provide colorful scribbles on the grayscale target image. The color information on the scribbles are then propagated to the rest of the target image using least-square optimization. Huang et al. \cite{Huang} develop an adaptive edge detection algorithm to reduce the color bleeding artifact around the region boundaries. Yatziv et al. \cite{Yatziv} colorize the pixels using a weighted combination of user¡¯ scribbles. Qu et al. \cite{Qu} and Luan et al. \cite{Luan} utilize the texture feature to reduce the amount of required scribbles.

\textbf{Example-based colorization} Unlike scribble-based colorization methods, the example-based methods transfer the color information from a reference image to the target grayscale image. The example-based colorization methods can be further divided into two categories according to the source of reference images:

(1) Colorization using user-supplied example(s). This type of methods requires the user to provide a suitable reference image. Inspired by image analogies \cite{Hertzmann} and the color transfer technique \cite{Reinhard}, Welsh et al. \cite{Welsh} employ the pixel intensity and neighborhood statistics to find a similar pixel in the reference image and then transfer the color of the matched pixel to the target pixel. It is later improved in \cite{irony} by taking into account the texture feature. Charpiat et al. \cite{charpiat} propose a global optimization algorithm to colorize a pixel. Gupta et al. \cite{gupta} develop an colorization method based on superpixel to improve the spatial coherency. These methods share the limitation that the colorization quality relies heavily on example image(s) provided by the user. However, there is not a standard criteria on the example image(s), thus finding a suitable reference image is a difficult task.

(2) Colorization using web-supplied example(s). To release the users' burden of finding a suitable image, Liu et al.\cite{liu} and Chia et al. \cite{chia} utilize the massive image data on the Internet. Liu et al.\cite{liu} compute an intrinsic image using a set of similar reference images collected from the Internet. This method is robust to illumination difference between the target and reference images, but it requires the images to contain identical object(s)/scene(s) for precise per-pixel registration between the reference images and the target grayscale image. It is unable to colorize the dynamic factors (\textit{e.g.} person, car) among the reference and target images, since these factors are excluded during the computation of the intrinsic image. As a result, it is limited to static scenes and the objects/scenes with a rigid shape (\textit{e.g.} famous landmarks). Chia et al. \cite{chia} propose an image filter framework to distill suitable reference images from the collected Internet images. It requires the user to provide semantic text label to search for suitable reference image on the Internet and human-segmentation cues for the foreground objects.

More recently, Deshpande et al. \cite{deshpande2015learning} propose a learning based framework that formulates this problem as a quadratic objective function. Histogram correction is applied to improve the initial colorization results. However, a suitable scene histogram is required in their refinement step. The other limitation is their low speed of colorization.

In contrast to the previous colorization methods, the proposed method is fully automatic by involving a large set of reference images from different scenes (\textit{e.g.}, coast, highway, field etc.) with various objects (\textit{e.g.}, tree, car, building etc.) and performs with artifact-free quality and high speed.

\section{Our Metric}
\label{sec:Proposed method}
An overview of the proposed colorization method is presented in Figure \ref{Fig:architecture}. Similar to the other learning based approaches, the proposed method has two major steps: (1) training a neural network assemble using a large set of example reference images; (2) using the learned neural network assemble to colorize a target grayscale image. These two steps are summarized in Algorithm \ref{alg:metric} and \ref{alg:metric_test}, respectively.

\begin{algorithm}
\caption{Image Colorization $-$ Training Step}
\begin{algorithmic}[1]
\REQUIRE Pairs of reference images: $\Lambda$ = $\{\vec{G},\vec{C}\}$.\\
\ENSURE A trained neural network assemble.\\
--------------------------------------------------------------------
\STATE Extract global descriptors of the reference images, group these images into different clusters adaptively and compute the semantic histogram of each cluster;\\
\STATE Compute feature descriptors $\vec{x}$ at sampled pixels in $\vec{G}$ and the corresponding chrominance values $\vec{y}$ in $\vec{C}$; \\
\STATE Construct a deep neural network for each cluster;\\
\STATE Train the deep neural networks using the training set $\Psi$ = $\{\vec{x}, \vec{y}\}$.\\
\end{algorithmic}
\label{alg:metric}
\end{algorithm}

\begin{algorithm}
\caption{Image Colorization $-$ Testing Step}
\begin{algorithmic}[1]
\REQUIRE A target grayscale image $I$ and the trained neural network assemble.\\
\ENSURE A corresponding color image: $\hat{I}$.\\
--------------------------------------------------------------------
\STATE Compute global descriptor and semantic histogram of $I$, then find its nearest cluster center and corresponding trained neural network; \\
\STATE Extract a feature descriptor at each pixel location in $I$;\\
\STATE Send feature descriptors extracted from $I$ to the trained neural network to obtain the corresponding chrominance values;\\
\STATE Refine the chrominance values to remove potential artifacts;\\
\STATE Combine the refined chrominance values and $I$ to obtain the color image $\hat{I}$.\\
\end{algorithmic}
\label{alg:metric_test}
\end{algorithm}

\subsection{A Deep Learning Model for Image Colorization}\label{sec:Deep learning model}
This section formulates image colorization as a regression problem and solves it using a regular deep neural network.

\subsubsection{Formulation} A deep neural network is a universal approximator that can represent arbitrarily complex continuous functions \cite{hornik}. Given a set of exemplars $\Lambda$ = $\{\vec{G},\vec{C}\}$, where $\vec{G}$ are grayscale images and $\vec{C}$ are corresponding color images respectively, our method is based on a premise: there exists a complex gray-to-color mapping function $\mathcal{F}$ that can map the features extracted at each pixel in $\vec{G}$ to the corresponding chrominance values in $\vec{C}$. We aim at learning such a mapping function from $\Lambda$ so that the conversion from a new gray image to color image can be achieved by using $\mathcal{F}$. In our model, the YUV color space is employed, since this color space minimizes the correlation between the three coordinate axes of the color space. For a pixel $p$ in $\vec{G}$ , the output of $\mathcal{F}$ is simply the U and V channels of the corresponding pixel in $\vec{C}$ and the input of $\mathcal{F}$ is the feature descriptors we compute at pixel $p$. The feature descriptors are introduced in detail in Sec. \ref{sec:Feature descriptor}. We reformulate the gray-to-color mapping function as $c_p$ = $\mathcal{F}(\Theta, x_p)$, where $x_p$ is the feature descriptor extracted at pixel $p$ and $c_p$ are the corresponding chrominance values. $\Theta$ are the parameters of the mapping function $\mathcal{F}$ to be learned from $\Lambda$.

We solve the following least squares minimization problem to learn the parameters $\Theta$:
\vspace{-1mm}
\begin{equation}
\argmin_{\Theta\subseteq\Upsilon}\sum_{p=1}^n\left\|\mathcal{F}(\Theta,x_p)-c_p\right\|^2
\vspace{-1mm}
\end{equation}
where $n$ is the total number of training pixels sampled from $\Lambda$ and $\Upsilon$ is the function space of $\mathcal{F}(\Theta,x_p)$.

\subsubsection{Architecture} Deep neural networks (DNNs) typically consist of one input layer, multiple hidden layers and one output layer. Generally, each layer can comprise various numbers of neurons. In our model, the number of neurons in the input layer is equal to the dimension of the feature descriptor extracted from each pixel location in a grayscale image and the output layer has two neurons which output the U and V channels of the corresponding color value, respectively. We perceptually set the number of neurons in the hidden layer to half of that in the input layer. Each neuron in the hidden or output layer is connected to all the neurons in the proceeding layer and each connection is associated with a weight. Let $o_j^l$ denote the output of the $j$-th neuron in the $l$-th layer. $o_j^l$ can be expressed as follows:

\vspace{-1mm}
\begin{equation}
o_j^l = f(w_{j0}^lb+\sum_{i>0}w_{ji}^lo_i^{l-1})
\vspace{-1mm}
\end{equation}

where $w_{ji}^l$ is the weight of the connection between the $j^{th}$ neuron in the $l^{th}$ layer and the $i^{th}$ neuron in the $(l-1)^{th}$ layer, the $b$ is the bias neuron which outputs value one constantly and $f(z)$ is an activation function which is typically nonlinear (\textit{e.g.}, tanh, sigmoid, ReLU\cite{krizhevsky}). The output of the neurons in the output layer is just the weighted combination of the outputs of the neurons in the proceeding layer. In our method, we utilize ReLU\cite{krizhevsky} as the activation function as it can speed up the convergence of the training process. The architecture of our neural network is presented in Figure \ref{Fig:architecture}.

We apply the classical error back-propagation algorithm to train the connected power of the neural network, and the weights of the connections between pairs of neurons in the trained neural network are the parameters $\Theta$ to be learned.

\subsection{Feature Descriptor}\label{sec:Feature descriptor}

Feature design is key to the success of the proposed colorization method. There are massive candidate image features that may affect the effectiveness of the trained model (\textit{e.g.} SIFT, SURF, Gabor, Location, Intensity histogram etc.). We conducted numerous experiments to test various features and kept only features that have practical impacts on the colorization results. We separate the adopted features into low-, mid- and high-level features. Let $x_p^L$, $x_p^M$, $x_p^H$ denote different-level feature descriptors extracted from a pixel location $p$, we concatenate these features to construct our feature descriptor $x_p$ = $\left\{x_p^L; x_p^M; x_p^H\right\}$. The adopted image features are discussed in detail in the following sections.

\subsubsection{Low-level Patch Feature}

\begin{figure}[!ht]
    \begin{center}
        \begin{tabular}{ccc}
        \includegraphics[width=\swidththree]{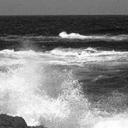}&
        \includegraphics[width=\swidththree]{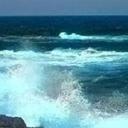}&
        \includegraphics[width=\swidththree]{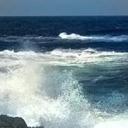}\\
        (a)Input &(b)-Patch feature &(c)+Patch feature
        \end{tabular}
    \end{center}
\caption{Evaluation of patch feature. (a) is the target grayscale image. (b) removes the low-level patch feature and (c) includes all the proposed features.}
\label{Fig:patch feature}
\end{figure}

Intuitively, there exist too many pixels with same luminance but fairly different chrominance in a color image, thus it's far from being enough to use only the luminance value to represent a pixel. In practice, different pixels typically have different neighbors, using a patch centered at a pixel $p$ tends to be more robust to distinguish pixel $p$ from other pixels in a grayscale image. Let $x_p^L$ denote the array containing the sequential grayscale values in a $7\times 7$ patch center at $p$, $x_p^L$ is used as the low-level feature descriptor in our framework. This feature performs better than traditional features like SIFT and DAISY at low-texture regions when used for image colorization. Figure \ref{Fig:patch feature} shows the impact of patch feature on our model. Note that our model will be insensitive to the intensity variation within a semantic region when the patch feature is missing (\textit{e.g.}, the entire sea region is assigned with one color in Figure \ref{Fig:patch feature}(b)).

\subsubsection{Mid-level DAISY Feature} \label{sec:Mid-level feature}

\begin{figure}[!ht]
    \begin{center}
        \begin{tabular}{cccc}
        \includegraphics[width=\swidthfour]{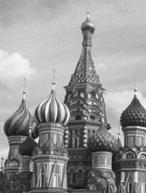}&
        \includegraphics[width=\swidthfour]{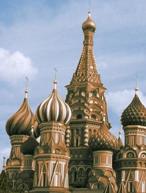}&
        \includegraphics[width=\swidthfour]{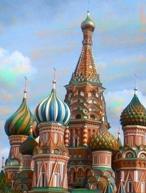}&
        \includegraphics[width=\swidthfour]{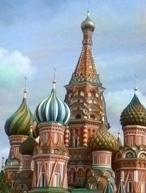}\\
        (a)Target&(b)-DAISY&(c)+DAISY&(d)Refined\\
        \includegraphics[width=\swidthfour]{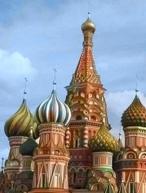}&
        \includegraphics[width=\swidthfour]{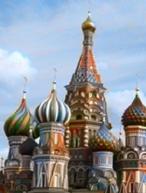}&
        \includegraphics[width=\swidthfour]{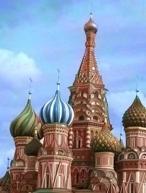}&
        \includegraphics[width=\swidthfour]{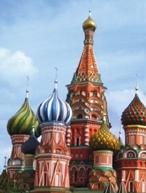}\\
        (e)Gupta \cite{gupta}&(f)Irony \cite{irony}&(g)Chia \cite{chia}&(h)Liu \cite{liu}\\
        \end{tabular}
    \end{center}
\caption{Evaluation of DAISY feature. (a) is the target gray scale image. (b) is our result without DAISY feature. (c) is our result after incorporating DAISY feature into our model. (d) is the final result after artifact removal (see Sec. \ref{sec:Refinement} for details). (e)-(h) presents results obtained with the state-of-the-art colorizations. Although the proposed method is fully-automatic, its performance is comparable to the state-of-the-art.}
\label{Fig:daisy feature}
\end{figure}

DAISY is a fast local descriptor for dense matching \cite{tola}. Unlike the low-level patch feature, DAISY can achieve a more accurate discriminative description of a local patch and thus can improve the colorization quality on complex scenarios. A DAISY descriptor is computed at a pixel location $p$ in a grayscale image and is denote as $x_p^M$. Figure \ref{Fig:daisy feature} demonstrates the performance with and without DAISY feature on a fine-structure object and presents the comparison with the state-of-the-art colorization methods. As can be seen, the adoption of DAISY feature in our model leads to a more detailed and accurate colorization result on complex regions. However, DAISY feature is not suitable for matching low-texture regions/objects and thus will reduce the performance around these regions as can be seen in Figure \ref{Fig:daisy feature}(c). A post-processing step will be introduced in Section \ref{sec:Refinement} to reduce the artifacts and its result is presented in Figure \ref{Fig:daisy feature}(d). Furthermore, we can see that our result is comparable to Liu et al. \cite{liu} (which requires a rigid-shape target object and identical reference objects) and Chia et al. \cite{chia} (which requires manual segmentation and identification of the foreground objects), although our method is fully-automatic.

\subsubsection{High-level Semantic Feature} \label{sec:High-level feature}
Patch and DAISY are low-level and mid-level features indicating the geometric structure of the neighbors of a pixel. The existing state-of-art methods typically employ such features to match pixels between the reference and target images. Recently, high-level properties of a image have demonstrated its importance and virtues in some fields (\textit{e.g.} image enhancement \cite{yan}, edge detection \cite{zheng}). Considering that the image colorization is typically a semantic-aware process, we extract a semantic feature at each pixel to express its category (\textit{e.g.} sky, sea, animal) in our model.

We adopt the state-of-art scene parsing algorithm \cite{long} to annotate each pixel with its category label, and obtain a semantic map for the input image. The semantic map is not accurate around region boundaries. As a result, it is smoothed using an efficient edge-preserving filter \cite{gastal} with the guidance of the original gray scale image. An N-dimension probability vector will be computed at each pixel location, where N is the total number of object categories and each element is the probability that the current pixel belongs to the corresponding category. This probability vector is used as the high-level descriptor denoted as $x_p^H$.

\begin{figure}[!ht]
    \begin{center}
        \begin{tabular}{ccc}
        \includegraphics[width=\swidththree]{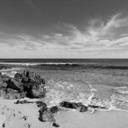}&
        \includegraphics[width=\swidththree]{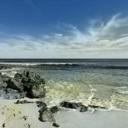}&
        \includegraphics[width=\swidththree]{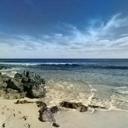}\\
        (a)Input&(b)Patch+DAISY&(c)+Semantic\\
        \end{tabular}
    \end{center}
\caption{Importance of semantic feature. (a) is the target grayscale image. (b) is the colorization result using patch and DAISY features only. (c) is the result using patch, DAISY and semantic features.}
\label{Fig:semantic feature}
\end{figure}

Figure \ref{Fig:semantic feature} shows that the colorization result may change significantly with and without the semantic feature. The adoption of semantic feature can significantly reduce matching/training ambiguities. For instance, if a pixel is detected to be inside a sky region, only sky color values resideing in the reference image database will be used. The colorization problem is thus simplified after integrating the semantic information and colorization result is visually much better as can be seen in Figure \ref{Fig:semantic feature}.

\subsubsection{Chrominance Refinement} \label{sec:Refinement}
The proposed method adopts the patch feature and DAISY feature, and we hope to use patch feature to describe low-texture simple regions and DAISY to describe fine-structure regions. However, we simply concatenate the two features instead of digging out a better combination. This will result in potential artifacts especially around the low-texture objects (\textit{e.g.}, sky, sea). This is because DAISY is vulnerable to these objects and presents a negative contribution.

\begin{figure}[!ht]
    \begin{center}
        \begin{tabular}{ccc}
        \includegraphics[width=\swidththree]{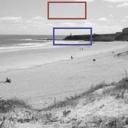}&
        \includegraphics[width=\swidththree]{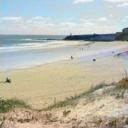}&
        \includegraphics[width=\swidththree]{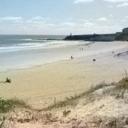}\\
        \includegraphics[width=\swidththree]{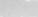}&
        \includegraphics[width=\swidththree]{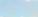}&
        \includegraphics[width=\swidththree]{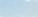}\\
        \includegraphics[width=\swidththree]{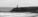}&
        \includegraphics[width=\swidththree]{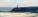}&
        \includegraphics[width=\swidththree]{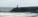}\\
        (a)Input&(b)Before&(c)After
        \end{tabular}
    \end{center}
  \caption{Chrominance refinement using joint bilateral filtering \cite{tog-04-Petschnigg}. From (a) to (c): target grayscale image, colorization results before and after chrominance refinement, respectively. Note that the artifacts in (b) are successfully removed from (c).}
 \label{Fig:Image Refinement}
\end{figure}

The artifacts around low-texture regions can be significantly reduced using joint bilateral filtering technique \cite{tog-04-Petschnigg}. It was first introduced to remove image noise of a no-flash image with the help of a noise-free flash image. Our problem is similar, the chrominance values obtained from the trained neural network is noisy (and thus results in visible artifacts) while the target grayscale image is noise-free. As a result, to ensure artifact-free quality, we apply joint bilateral filtering to smooth/refine the chrominance values (computed by the trained neural network) with the target grayscale image as the guidance. Figure \ref{Fig:Image Refinement} presents the result before and after chrominance refinement. Note that most of the visible artifacts can be successfully removed.

\begin{figure*}[!ht]
    \begin{center}
{\small
        \begin{tabular}{cccccccc}
        \includegraphics[width=\swidthseven]{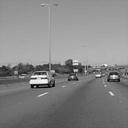}&
        \includegraphics[width=\swidthseven]{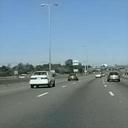}&
        \includegraphics[width=\swidthseven]{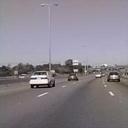}&
        \includegraphics[width=\swidthseven]{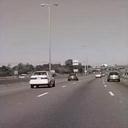}&
        \includegraphics[width=\swidthseven]{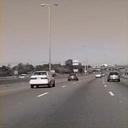}&
        \includegraphics[width=\swidthseven]{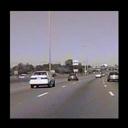}&
        \includegraphics[width=\swidthseven]{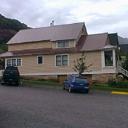}\\
        & 33dB & 22dB & 21dB & 20dB & 11dB & \\
        (a) Input image&(b) Our method&(c) Gupta et al.&
        (d) Irony et al.&(e) Welsh et al.&(f) Charpiat et al.&
        (g) Reference image\\
        &     & \cite{gupta} & \cite{irony} & \cite{Welsh} & \cite{charpiat} &  \\
        \end{tabular}
}
    \end{center}
  \caption{Comparison with the state-of-art colorization methods \cite{charpiat, gupta, irony, Welsh}. (c)-(f) use (g) as the reference image, while the proposed method adopts a large reference image dataset. The reference image contains similar objects as the target grayscale image (e.g., road, trees, building, cars). It is seen that the performance of the state-of-art colorization methods is lower than the proposed method when the reference image is not ``optimal''. The segmentation masks used by \cite{irony} are computed by mean shift algorithm \cite{comaniciu}. The PSNR values computed from the colorization results and ground truth are presented under the colorized images.}
 \label{Fig:comparison1}
\end{figure*}

\subsection{Adaptive Image Clustering}\label{sec: Clustering}
This section presents an adaptive image clustering technique and demonstrates its effectiveness in improving the colorization performance.

The proposed DNN trained from a large reference image set that contains various scenes performs well in most cases. However, visible artifacts still appear, especially on the objects with large color variances (e.g. building, plants etc.). One reason is that the receptive field of the DNN is limited on local patch, which causes large training ambiguities especially when large training set is utilized. Intuitively, the global image descriptor (\textit{e.g.} gist \cite{gist}, intensity histogram etc.) is able to reflect the scene category (\textit{e.g}. coast, highway, city etc.) with the robustness to local noise, and there are typically smaller color variances within one scene than mixed scenes. Thus the global information is useful to reduce the matching/training ambiguities and improve the colorization accuracy. \cite{cheng2015deep} reveals that feeding the global descriptor into DNN directly would produce an unnatural colorization result. In the present work, we incorporate the global information by an image clustering method. Inspired by \cite{ren2015image} which adopts an adaptive pixel clustering algorithm and trains a regressor assemble to model the light transport, we utilize a similar strategy to split the reference images into different scenes, for each of which a DNN is trained.

As illustrated in Algorithm \ref{alg:clustering}, the reference images are clustered adaptively on different layers by standard k-means clustering algorithm. After completing the training of DNN for cluster $i$ on layer $l$, we measure the training error $E(I_{(i, l)})$ for each reference image $I_{(i, l)}$ as the negative Peak Signal-to-Noise Ratio (PSNR) computed from the colorization result $\hat{I}_{(i, l)}$ and the ground truth image. If $E(I_{(i, l)})$ is lower than a threshold $\varepsilon$, $I_{(i, l)}$ will be removed from the reference image set $\Lambda_{(i, l)}$. As a result, the top layer contains all reference images while the lower layer comprises fewer images.

To ensure a sufficient number of samples for training a single DNN, the number of clusters on the next lower layer is determined by the size of $\Lambda$ as well as the minimal number of reference images required for training a single DNN (denoted as $\mu$). Similar to \cite{ren2015image}, we compute $\mu$ by the following equation according to \cite{turmon1995sample}:

\vspace{-1mm}
\begin{equation}
\mu = \frac{\alpha N_w}{N_s}
\vspace{-1mm}
\end{equation}

where $\alpha$ is a constant scale factor, $N_w$ is the total number of weights in a single DNN, and $N_s$ is the number of samples from one reference image.

\begin{algorithm}
\caption{Adaptive Image Clustering}
\begin{algorithmic}[1]
\REQUIRE Pairs of reference images: $\Lambda$ = $\{\vec{G},\vec{C}\}$; Error threshold: $\varepsilon$; Minimal number of reference images required for training one DNN: $\mu$; Initial number of clusters on the top layer: $N^0$ \\
\ENSURE Trained DNN assemble $\Phi$; Hierarchy cluster assemble $\Omega$.\\
-------------------------------------------------------------------- \\
\STATE Extract global descriptors of reference images $\Lambda$;
\STATE $l := 0$; // the top layer
\WHILE {$size$$($$\Lambda$$)$ $>=$ $\mu$}
\STATE Group $\Lambda$ into $N^l$ clusters $\Omega_{(1...N^l, l)}$ on layer $l$;
\STATE Compute semantic histogram for each cluster $\Omega_{(i, l)}$;
\STATE Train a DNN $\Phi_{(i, l)}$ for each cluster $i$ on layer $l$ using the reference images $\Lambda_{(i, l)}$ = $\{\vec{G}_{(i, l)},\vec{C}_{(i, l)}\}$£»
\FOR {\textit{each reference image $I_{(i, l)}$ in $\Lambda_{(i, l)}$}}
\STATE Measure training error $E(I)$ ;
\IF {$E(I)$ $<=$ $\varepsilon$}
\STATE Remove $I$ from $\Lambda_{(i, l)}$;
\ENDIF
\STATE $l := l + 1$, $N^l$ $:=$ $size$$($$\Lambda_{(i, l)}$$)$ $/$ $\mu$;
\ENDFOR
\ENDWHILE
\end{algorithmic}
\label{alg:clustering}
\end{algorithm}

\subsubsection{Semantic Histogram}
\begin{figure}[!ht]
    \begin{center}
        \begin{tabular}{cccc}
        \includegraphics[width=\swidthfour]{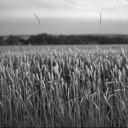}&
        \includegraphics[width=\swidthfour]{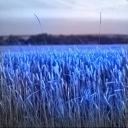}&
        \includegraphics[width=\swidthfour]{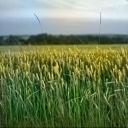}&
        \includegraphics[width=\swidthfour]{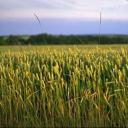}\\
        (a) Input &(b) -Semantic&(c) +Semantic&(d) GT\\
        &               histogram &   histogram & \\
        \end{tabular}
    \end{center}
  \caption{Evaluation of semantic histogram. (a) is the input image. (b) is the colorization result when the semantic histogram is not used in nearest cluster searching. (c) is result after incorporating semantic histogram. (d) is the ground truth.}
 \label{Fig:stchist}
\end{figure}

After scene-wise DNNs are trained, a straightforward colorization strategy is to find the nearest cluster for a target image and use the corresponding trained DNN to colorize it. However, it is very likely that the reference images in the searched cluster are globally similar but semantically different from the target images. For example, the nearest cluster for Figure \ref{Fig:stchist}(a) searched using only global image feature belongs to the ``building'' scene, which causes an unnatural colorization result, as shown in Figure \ref{Fig:stchist} (b).

To address this problem, we incorporate the semantic histogram to search for the globally and semantically similar cluster. The number of bins is equal to the predefined object categories. And each bin that represents the percentage of pixels belongs to a certain object. In test phrase, we first search for the top-$k$ nearest clusters by the Euclidean distance of global descriptors between the clusters and the target image, then find out the nearest cluster by the cosine similarity of semantic histogram within the initial $k$ clusters. Figure \ref{Fig:stchist} shows the performance could change significantly  with and without semantic histogram.

\subsubsection{The Evaluation of Image Clustering}
\begin{figure}[!ht]
    \begin{center}
        \begin{tabular}{c}
        \includegraphics[width=\swidthone]{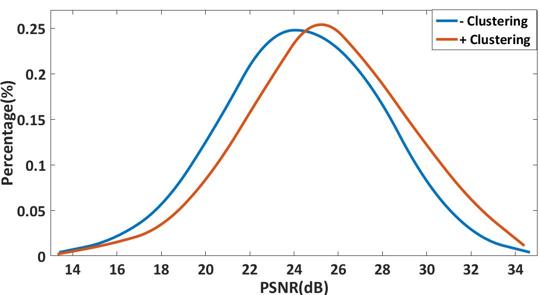}\\
        \end{tabular}
    \end{center}
  \caption{The PSNR distribution with/without image clustering. It is seen that the proposed image clustering technique can improve the colorization accuracy.}
 \label{Fig:psnr1}
\end{figure}

\begin{figure}[!ht]
    \begin{center}
        \begin{tabular}{cccc}
        \includegraphics[width=\swidthfour]{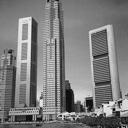}&
        \includegraphics[width=\swidthfour]{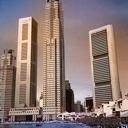}&
        \includegraphics[width=\swidthfour]{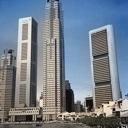}&
        \includegraphics[width=\swidthfour]{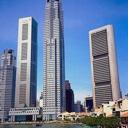}\\
        &16dB&23dB&\\
        \includegraphics[width=\swidthfour]{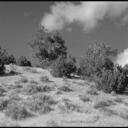}&
        \includegraphics[width=\swidthfour]{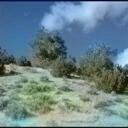}&
        \includegraphics[width=\swidthfour]{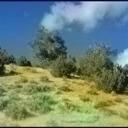}&
        \includegraphics[width=\swidthfour]{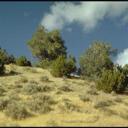}\\
        &20dB&23dB&\\
        \includegraphics[width=\swidthfour]{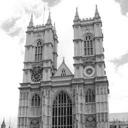}&
        \includegraphics[width=\swidthfour]{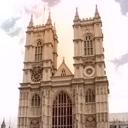}&
        \includegraphics[width=\swidthfour]{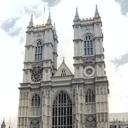}&
        \includegraphics[width=\swidthfour]{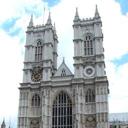}\\
        &22dB&31dB&\\
        (a) Input &(b) -Clustering &(c) +Clustering &(d) GT\\
        \end{tabular}
    \end{center}
  \caption{Evaluation of the adaptive image clustering technique. (a) are the input images. (b) are the colorization results without image clustering. (c) are the results of the proposed method that utilizes image clustering. (d) are the ground truth of (a). The PSNR values computed from the colorization results and the ground truth are presented under the colorized images. }
 \label{Fig:dnn_vs_dnns}
\end{figure}

Figure \ref{Fig:psnr1} presents the PSNR distribution of 1519 test images with/without image clustering. Figure \ref{Fig:dnn_vs_dnns} shows the qualitative comparisons. It is seen that the proposed image clustering technique can improve the colorization accuracy and reduce the visible artifacts significantly, especially for the objects with large color variances (\textit{e.g.} building, plant etc.)

\subsection{Difference from the State-of-the-art Colorization Methods}

\begin{figure}[!ht]
    \begin{center}
{\small
        \begin{tabular}{cccc}
        \includegraphics[width=\swidththree]{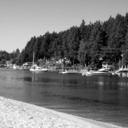}&
        \includegraphics[width=\swidththree]{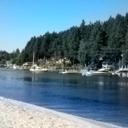}&
        \includegraphics[width=\swidththree]{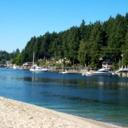}\\
        (a) Input & (b) Proposed & (c) Ground truth\\
                  &         26dB &  \\
        \includegraphics[width=\swidththree]{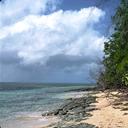}&
        \includegraphics[width=\swidththree]{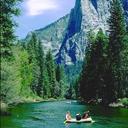}&
        \includegraphics[width=\swidththree]{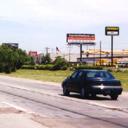}\\
        (d) Reference 1& (e) Reference 2& (f)  Reference 3 \\
        \includegraphics[width=\swidththree]{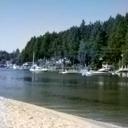}&
        \includegraphics[width=\swidththree]{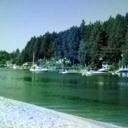}&
        \includegraphics[width=\swidththree]{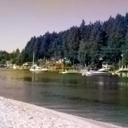}\\
        (g) \cite{gupta}+(d) & (h) \cite{gupta}+(e) &  (i) \cite{gupta}+(f) \\
               21dB  &         20dB  &         17dB \\
        \end{tabular}
}
    \end{center}
  \caption{The high dependence on a suitable reference image of Gupta et al. \cite{gupta}. (a) is the input grayscale image. (b) is the color image obtained by the proposed method which is visually more accurate. (c) is the ground truth of (a). (d) is the first reference image for \cite{gupta}. It has a similar scene as the (a). (e) is the second reference image that also has similar scene but lacks 'beach' object. (f) is the last reference image that is complete different from (a). The color images obtained from \cite{gupta} w.r.t. the reference images in (d)-(f) are presented in (g)-(i), respectively. The PSNR values computed from the colorization results and the ground truth are presented under the colorized images.}
 \label{Fig:High ralience}
\end{figure}

The previous algorithms \cite{charpiat, chia, gupta, irony,Welsh} typically use one similar reference image or a set of similar reference images from which transfer color values to the target gray image. \cite{gupta} is the state-of-art example-based method as it outperforms others in performance and application scope. However, its performance highly depends on given reference image as demonstrated in Figure \ref{Fig:High ralience}. \cite{gupta} can obtain a very good colorization result using a reference image containing identical object(s) as the target grayscale image. However, when the reference image is different from the target, its performance is quite low as shown in Figure \ref{Fig:High ralience} (h)-(i). To minimize the high dependence on a suitable reference image, our method utilizes a large reference image database. It ``finds'' the most similar pixel from the database and ``transfers'' its color to the target pixel. This is why our approach is robust to different grayscale target images.

\begin{figure}[!ht]
    \begin{center}
{\small
        \begin{tabular}{cc}
        \includegraphics[width=\swidthfour]{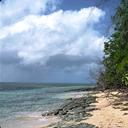}
        \includegraphics[width=\swidthfour]{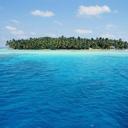}
        \includegraphics[width=\swidthfour]{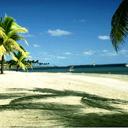}&
        \includegraphics[width=\swidthfour]{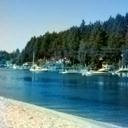}\\
        (a) Reference image set 1&(d) \cite{gupta}+(a) (22dB)\\
        \includegraphics[width=\swidthfour]{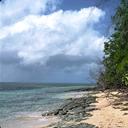}
        \includegraphics[width=\swidthfour]{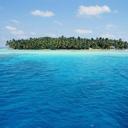}
        \includegraphics[width=\swidthfour]{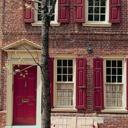}&
        \includegraphics[width=\swidthfour]{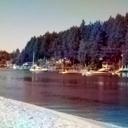}\\
        (b) Reference image set 2&(e) \cite{gupta}+(b) (17dB)\\
        \includegraphics[width=\swidthfour]{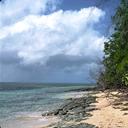}
        \includegraphics[width=\swidthfour]{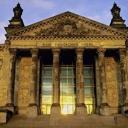}
        \includegraphics[width=\swidthfour]{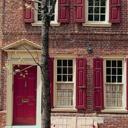}&
        \includegraphics[width=\swidthfour]{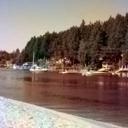}\\
        (c) Reference image set 3&(f) \cite{gupta}+(c) (16dB)\\
        \end{tabular}
}
    \end{center}
  \caption{Gupta et al.\cite{gupta} with multiple reference images. The target grayscale image is the same as Figure \ref{Fig:High ralience}(a). (a)-(c) are different reference images and (d)-(f) are the corresponding colorization results. Note that the best performance can be achieved when sufficient similar reference images are used.}
 \label{Fig:Multi-refer}
\end{figure}

Intuitively, one reference image cannot comprise all suitable correspondences for pixels in the target grayscale image. This is why the performance of \cite{gupta} highly depends on a suitable reference image. As shown in Figure \ref{Fig:Multi-refer}, using a couple of similar reference images could improve their colorization result. However, when the reference images contain multiple objects (\textit{e.g.} door, window, building etc.), their colorization result becomes unnatural, although some of the reference images are similar to the target. This is due to the significant amount of noise residing in feature matching (between the reference images and the target image). For instance, we noticed that the lake in Figure \ref{Fig:High ralience}(a) was matched to the door in Figure \ref{Fig:Multi-refer}(e)), and the sky was matched to the building in Figure \ref{Fig:Multi-refer}(f).

Experiments demonstrate that deep learning techniques are well-suited for a large reference image database. The deep neural network helps to combine the various features of a pixel and computes the corresponding chrominance values. Additionally, the state-of-the-art methods are very slow because they have to find the most similar pixels (or super-pixels) from massive candidates. In comparison, the deep neural network is tailored to massive data. Although the training of neural networks is slow especially when the database is large, colorizing a 256$\times$256 grayscale image using the trained neural network assemble takes only 6.780 seconds in Matlab.

More recently, Deshpande et al. \cite{deshpande2015learning} propose an automatic colorization framework. Similar to our method, \cite{deshpande2015learning} solves this problem by minimizing a quadratic objective function, and also proposes an post-processing technique to improve their colorization performance. The main differences lie in the following aspects:

\begin{enumerate}
  \item The proposed deep neural networks learn the mapping function automatically, so that we need not design the objective function carefully by hand or search for massive hyper-parameters like \cite{deshpande2015learning};
  \item To achieve good performances, \cite{deshpande2015learning} requires a suitable scene histogram in their refinement step. Their best colorization results are typically obtained by using the ground-truth scene histogram. By contrast, no such spatial prior is required for the proposed method.
  \item The proposed model colorizes a target image at a much higher speed than \cite{deshpande2015learning}. It takes only 6.780 seconds to colorize a 256$\times$256 using the proposed model while \cite{deshpande2015learning} requires 251.709 seconds and more time to adjust the histograms in their refinement step.
\end{enumerate}

\section{Experimental Results}
\label{sec:Experiments}

The proposed colorization neural network assemble is trained on 2344 images from the SIFT Flow database (a subset of SUN Attribution database \cite{sun}). We evaluate the proposed model on 1519 images from Sun database \cite{xiao2010sun}. Each image is segmented into a number of object regions and a total of 33 object categories \footnote{There is one error in \cite{cheng2015deep}. Only 33 (instead of 47) object categories were used in \cite{cheng2015deep}.} are used (\textit{e.g.} building, car, sea etc.). The neural network has an input layer, three hidden layers and one output layer. According to our experiments, using more hidden layers cannot further improve the colorization results. A 49-dimension ($7\times7$) patch feature, a 32-dimension DAISY feature \cite{tola} (4 locations and 8 orientations) and a 33-dimension semantic feature are extracted at each pixel location. Thus, there are a total of 114 neurons in the input layer. This paper perceptually sets the number of neurons in the hidden layer to half of that in the input layer and 2 neurons in the output layer (which correspond to the chrominance values). The parameters $\varepsilon$, $\mu$, $N^0$ for the proposed adaptive image clustering are set to -26dB, 80 and 24 respectively. We use gist feature \cite{gist} as the global image descriptor in our experiment.

\subsection{Scene Parsing on Grayscale Image}

We retrained the semantic segmentation model proposed by \cite{long} using the grayscale version of images from SIFT Flow dataset and evaluated the trained model on the standard 200 test images. As shown in Table \ref{table:stc}, \cite{long} outperforms other algorithms \cite{liu2011sift, tighe2013finding, farabet2013learning, pinheiro2014recurrent} in terms of pixel accuracy, whether trained by color or grayscale images. It also proves that the color information is useful for scene parsing, as the best performance is achieved by training \cite{long} using color images. We verify that the retrained model of \cite{long} on grayscale images is sufficient enough for our colorization work.

\begin{table}
\begin{center}
\renewcommand\arraystretch{1.2}
\caption{Performance comparison of semantic segmentation model. The first column lists the existing state-of-art scene parsing algorithms, and the second column shows the version of training and test images. The last column presents the standard metric (i.e. pixel accuracy) for evaluation.}
\label{table:stc}
\begin{tabular}{m{70pt}<{\centering} m{50pt}<{\centering} m{50pt}<{\centering}}
 \toprule
\textbf{Methods} & \textbf{Image Version} & \textbf{\textit{Pixel Acc.}}
\\ \midrule
Long et al. \cite{long} & color& 85.2\\
Long et al. \cite{long} & grayscale& 78.9\\
Liu et al. \cite{liu2011sift} & color& 76.7\\
Tighe et al.\cite{tighe2013finding} 1 & color& 75.6\\
Tighe et al.\cite{tighe2013finding} 2 & color& 78.6\\
Farabet et al. \cite{farabet2013learning} 1 & color& 72.3\\
Farabet et al. \cite{farabet2013learning} 2 & color& 78.5\\
Pinheiro et al. \cite{pinheiro2014recurrent} & color& 77.7\\
\bottomrule
\end{tabular}
\end{center}
\end{table}

\subsection{Comparisons with State-of-the-Arts}
Figure \ref{Fig:comparison1} compares our colorization results with the state-of-the-art colorization methods \cite{charpiat,gupta,irony,Welsh}.
The performance of these colorization methods is very high when an ``optimal'' reference image is used (e.g., containing the same objects as the target grayscale image), as shown in \cite{charpiat,gupta,irony,Welsh}. However, the performance may drop significantly when the reference image is only similar to the target grayscale image. The proposed method does not have this limitation due to the use of a large reference image database as shown in Figure \ref{Fig:architecture} (a).

\begin{figure}[!ht]
    \begin{center}
        \begin{tabular}{ccccc}
                \includegraphics[width=\swidthfive]{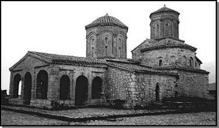}&
        \includegraphics[width=\swidthfive]{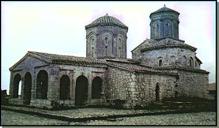}&
        \includegraphics[width=\swidthfive]{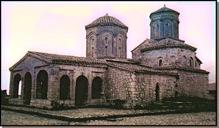}&
        \includegraphics[width=\swidthfive]{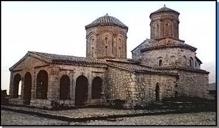}&
        \includegraphics[width=\swidthfive]{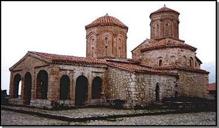}\\
        \includegraphics[width=\swidthfive]{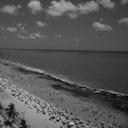}&
        \includegraphics[width=\swidthfive]{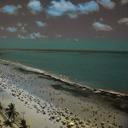}&
        \includegraphics[width=\swidthfive]{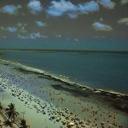}&
        \includegraphics[width=\swidthfive]{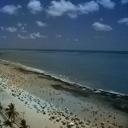}&
        \includegraphics[width=\swidthfive]{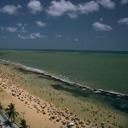}\\
        \includegraphics[width=\swidthfive]{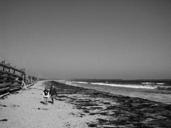}&
        \includegraphics[width=\swidthfive]{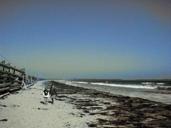}&
        \includegraphics[width=\swidthfive]{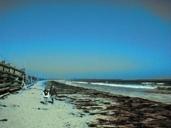}&
        \includegraphics[width=\swidthfive]{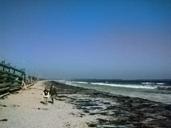}&
        \includegraphics[width=\swidthfive]{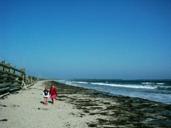}\\
                \includegraphics[width=\swidthfive]{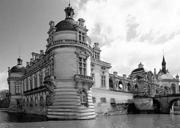}&
        \includegraphics[width=\swidthfive]{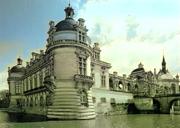}&
        \includegraphics[width=\swidthfive]{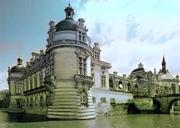}&
        \includegraphics[width=\swidthfive]{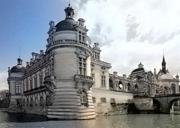}&
        \includegraphics[width=\swidthfive]{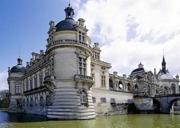}\\
        \includegraphics[width=\swidthfive]{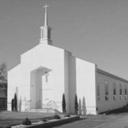}&
        \includegraphics[width=\swidthfive]{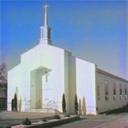}&
        \includegraphics[width=\swidthfive]{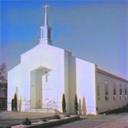}&
        \includegraphics[width=\swidthfive]{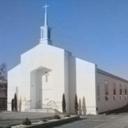}&
        \includegraphics[width=\swidthfive]{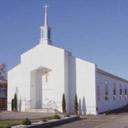}\\
                \includegraphics[width=\swidthfive]{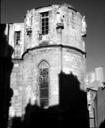}&
        \includegraphics[width=\swidthfive]{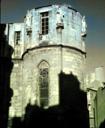}&
        \includegraphics[width=\swidthfive]{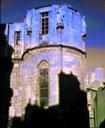}&
        \includegraphics[width=\swidthfive]{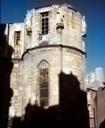}&
        \includegraphics[width=\swidthfive]{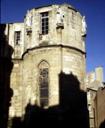}\\
        (a) Input &(b) \cite{deshpande2015learning} &(c) \cite{deshpande2015learning}&(d) Ours &(e) GT\\
                  &     +Mean Hist                   &          +GT Hist               &         &      \\
        \end{tabular}
    \end{center}
  \caption{Comparison with Deshpande et al. \cite{deshpande2015learning}. (a) are the input grayscale images. (b) are the results generated by \cite{deshpande2015learning} using mean color histogram computed from reference images. (c) are the results of \cite{deshpande2015learning} using ground-truth color histogram of (a). (d) are the results of the proposed model. (e) are the ground truth of (a).}
 \label{Fig:comparison2}
\end{figure}

Figure \ref{Fig:comparison2} shows the comparison with \cite{deshpande2015learning}. It is seen that \cite{deshpande2015learning} performs well when a suitable scene histogram is used in their refinement step, but visible artifacts still appear frequently. By contrast, the proposed method generates more natural colorization results with higher spatial coherency and fewer artifacts, and no spatial priors are required.

\subsection{Colorization in Different Global Styles}
\begin{figure}[!ht]
    \begin{center}
        \begin{tabular}{cccc}
        \includegraphics[width=\swidthfour]{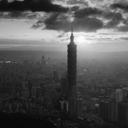}&
        \includegraphics[width=\swidthfour]{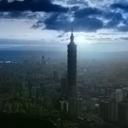}&
        \includegraphics[width=\swidthfour]{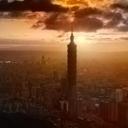}&
        \includegraphics[width=\swidthfour]{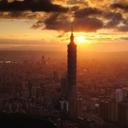}\\
        \includegraphics[width=\swidthfour]{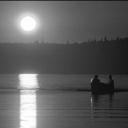}&
        \includegraphics[width=\swidthfour]{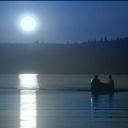}&
        \includegraphics[width=\swidthfour]{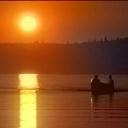}&
        \includegraphics[width=\swidthfour]{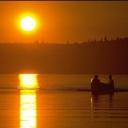}\\
        (a) Input &(b) Daytime&(c) Dusk &(d) GT\\
        \end{tabular}
    \end{center}
  \caption{Colorization in daytime/dusk style. (a) is the input image. (b) is the colorization results in daytime style. (c) is results in dusk style. (d) is the ground truth.}
 \label{Fig:style}
\end{figure}

One problem of \cite{cheng2015deep} is that it colorizes the target grayscale image in one global style. For example, as shown in Figure \ref{Fig:style} (b), all grayscale images are colorized in a daytime style automatically. Although these colorization results are visually reasonable, it is possible that the user has special requirements on the colorization style (e.g. dusk). However, given a grayscale image, it is very challenging to recognize whether it belongs to daytime or dusk even by human eyes, which makes it hard to generate more than one colorization styles using an uniform neural network. An alternative is to train a specific neural network for the required global style. Our experiments show that the proposed model is flexible to learn different global styles, as shown in Figure \ref{Fig:style}.

\subsection{Running Time}
The proposed model is able to process images of any resolutions at a high speed. Table \ref{table:runtime} shows the average running time on images of different resolutions on a computer equipped with \textit{Intel$^\circledR$ Xeon$^\circledR$ @ 2.30GHz} CPU, along with the comparison to \cite{deshpande2015learning}\footnote{We use the source code released by \cite{deshpande2015learning} in our experiment. \url{http://vision.cs.illinois.edu/projects/lscolor}}. It is seen that the proposed model is much faster than \cite{deshpande2015learning}, and our running time increases nearly linearly with the image resolution.

\begin{table}
\begin{center}
\renewcommand\arraystretch{1.2}
\caption{Running Time (seconds) on images of different resolutions, and comparison to Deshpande et al. \cite{deshpande2015learning}. Note that we only compare with \cite{deshpande2015learning} here, since both the proposed method and \cite{deshpande2015learning} are fully-automatic while other colorization methods \cite{Huang,Levin,Luan, Qu, Yatziv, charpiat, chia, gupta, irony, liu, Welsh} typically require efforts from the user, which makes it hard to measure their running time.}
\label{table:runtime}
\begin{tabular}{m{50pt}<{\centering} m{40pt}<{\centering} m{40pt}<{\centering} m{40pt}<{\centering} }
 \toprule
\textbf{Image Size} & 256$\times$256  & 512$\times$512 & 1024$\times$1024
\\ \midrule
\textbf{Proposed} &  6.780 & 17.413 & 63.142\\
\textbf{\cite{deshpande2015learning}} &  251.709 & 712.149 & 5789.075\\
\bottomrule
\end{tabular}
\end{center}
\end{table}

\subsection{More Colorization Results}
\begin{figure*}[!ht]
    \begin{center}
{\small
        \begin{tabular}{ccccccccc}
        \rotatebox{90}{Input}&
        \includegraphics[width=\swidtheight]{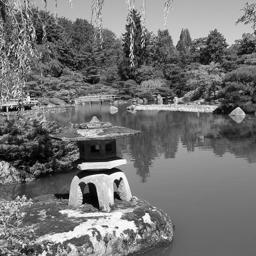}&
        \includegraphics[width=\swidtheight]{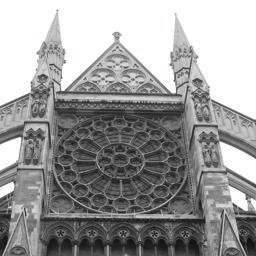}&
        \includegraphics[width=\swidtheight]{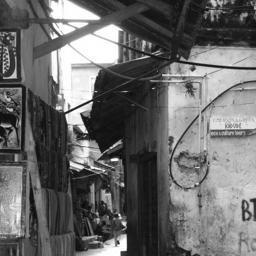}&
        \includegraphics[width=\swidtheight]{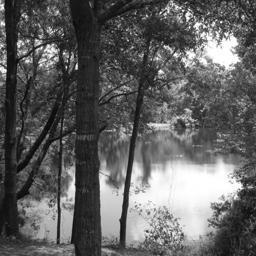}&
        \includegraphics[width=\swidtheight]{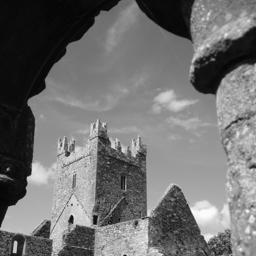}&
        \includegraphics[width=\swidtheight]{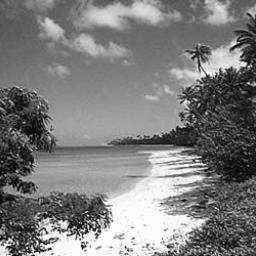}&
        \includegraphics[width=\swidtheight]{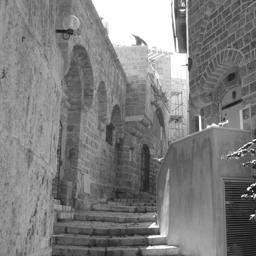}&
                \includegraphics[width=\swidtheight]{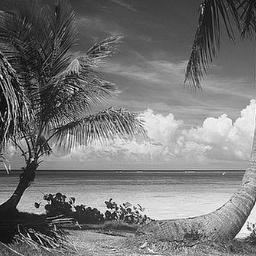}\\
        \rotatebox{90}{Proposed}&
        \includegraphics[width=\swidtheight]{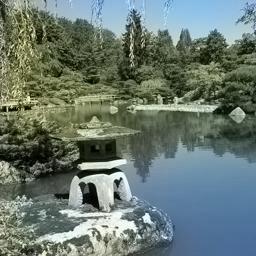}&
        \includegraphics[width=\swidtheight]{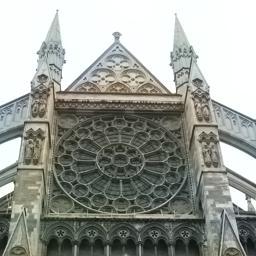}&
        \includegraphics[width=\swidtheight]{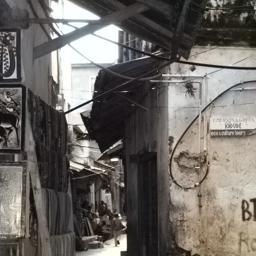}&
        \includegraphics[width=\swidtheight]{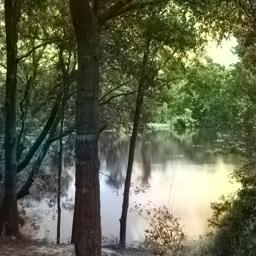}&
        \includegraphics[width=\swidtheight]{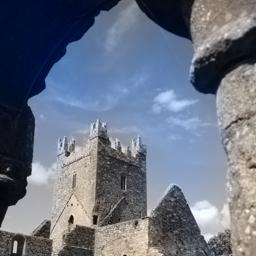}&
        \includegraphics[width=\swidtheight]{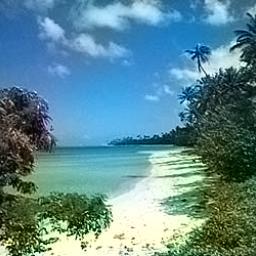}&
        \includegraphics[width=\swidtheight]{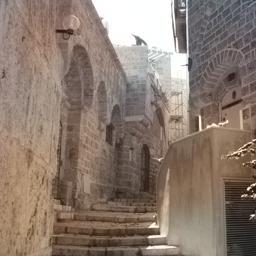}&
				\includegraphics[width=\swidtheight]{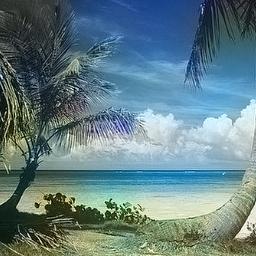}\\
        &25dB&33dB&29dB&25dB&24dB&23dB&24dB&25dB\\
        \rotatebox{90}{Groundtruth}&
		\includegraphics[width=\swidtheight]{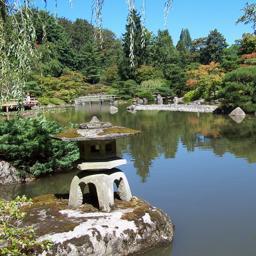}&
        \includegraphics[width=\swidtheight]{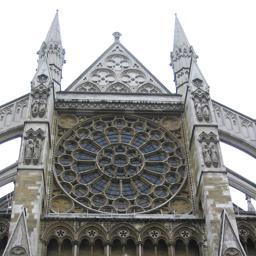}&
        \includegraphics[width=\swidtheight]{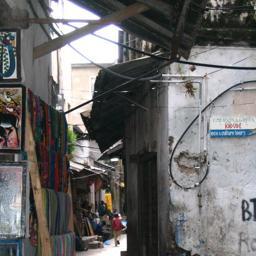}&
        \includegraphics[width=\swidtheight]{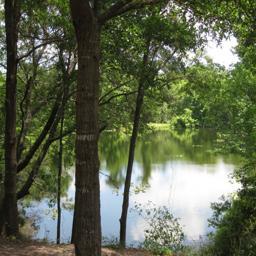}&
        \includegraphics[width=\swidtheight]{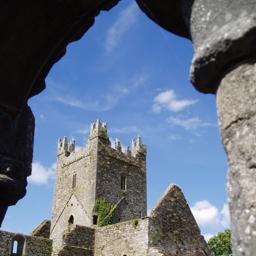}&
        \includegraphics[width=\swidtheight]{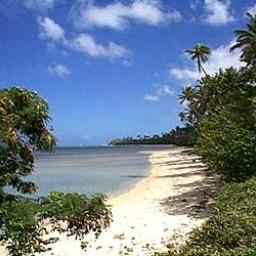}&
        \includegraphics[width=\swidtheight]{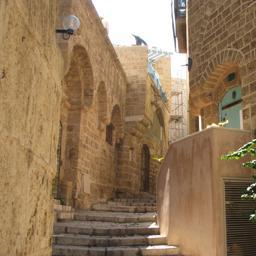}&
                \includegraphics[width=\swidtheight]{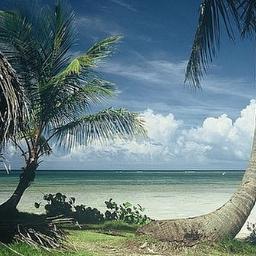}\\
        \rotatebox{90}{Input}&
        \includegraphics[width=\swidtheight]{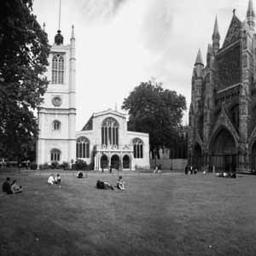}&
        \includegraphics[width=\swidtheight]{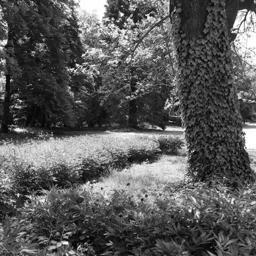}&
        \includegraphics[width=\swidtheight]{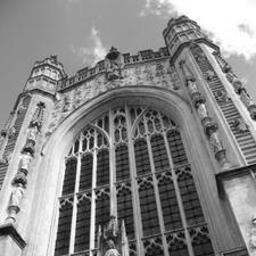}&
        \includegraphics[width=\swidtheight]{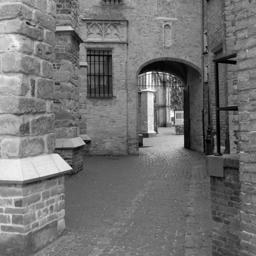}&
        \includegraphics[width=\swidtheight]{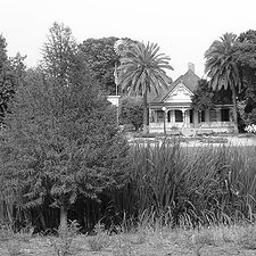}&
        \includegraphics[width=\swidtheight]{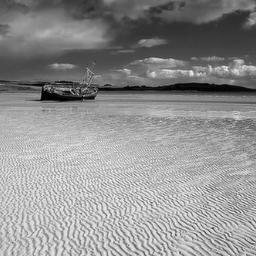}&
        \includegraphics[width=\swidtheight]{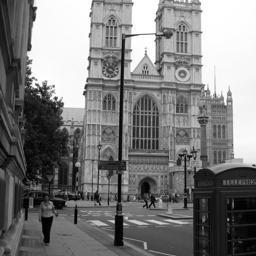}&
				\includegraphics[width=\swidtheight]{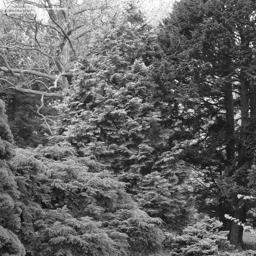}\\
        \rotatebox{90}{Proposed}&
        \includegraphics[width=\swidtheight]{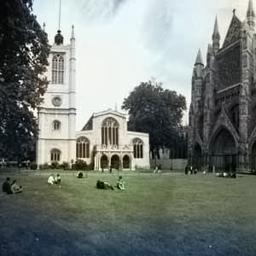}&
        \includegraphics[width=\swidtheight]{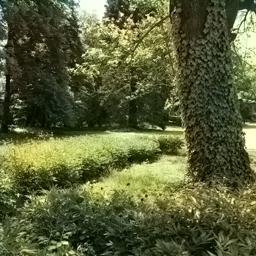}&
        \includegraphics[width=\swidtheight]{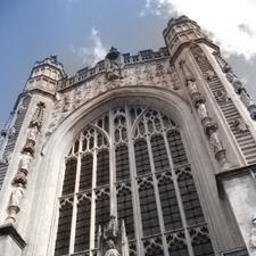}&
        \includegraphics[width=\swidtheight]{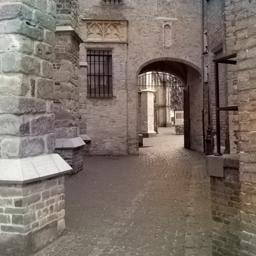}&
        \includegraphics[width=\swidtheight]{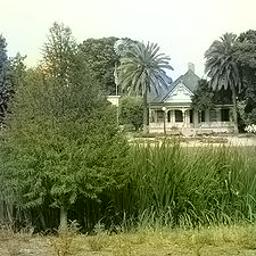}&
        \includegraphics[width=\swidtheight]{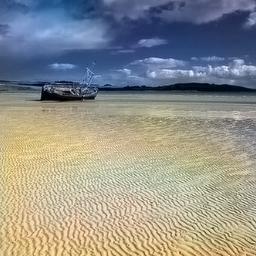}&
        \includegraphics[width=\swidtheight]{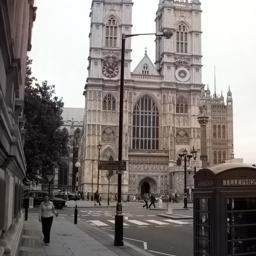}&
				\includegraphics[width=\swidtheight]{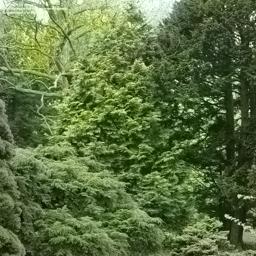}\\
        &25dB&29dB&24dB&27dB&27dB&22dB&31dB&26dB\\
        \rotatebox{90}{Groundtruth}&
		\includegraphics[width=\swidtheight]{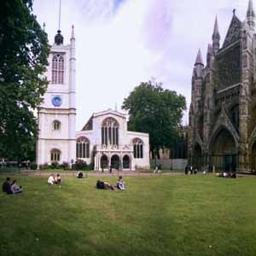}&
        \includegraphics[width=\swidtheight]{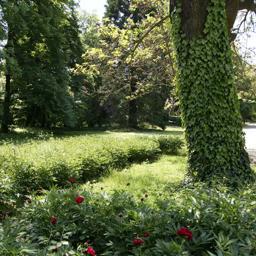}&
        \includegraphics[width=\swidtheight]{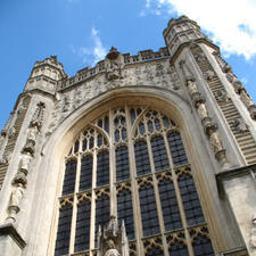}&
        \includegraphics[width=\swidtheight]{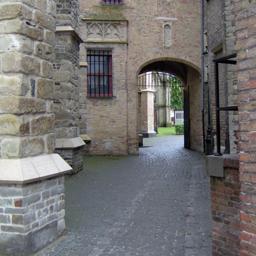}&
        \includegraphics[width=\swidtheight]{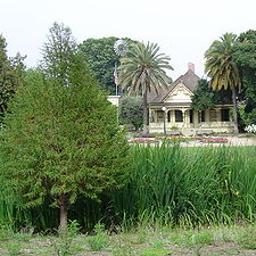}&
        \includegraphics[width=\swidtheight]{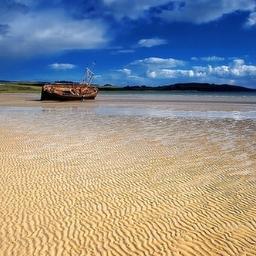}&
        \includegraphics[width=\swidtheight]{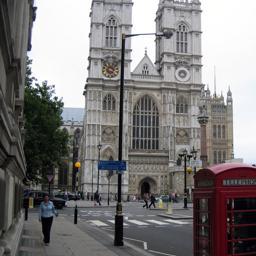}&
				\includegraphics[width=\swidtheight]{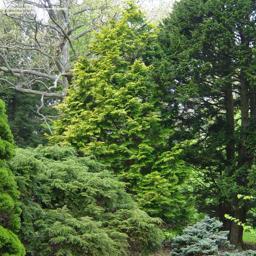}\\
        \rotatebox{90}{Input}&
        \includegraphics[width=\swidtheight]{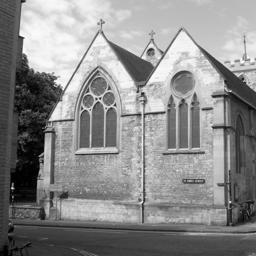}&
        \includegraphics[width=\swidtheight]{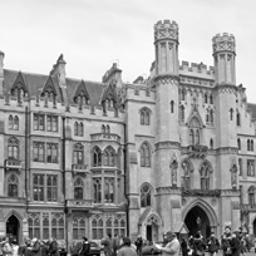}&
        \includegraphics[width=\swidtheight]{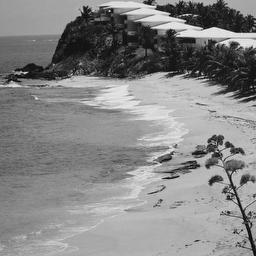}&
        \includegraphics[width=\swidtheight]{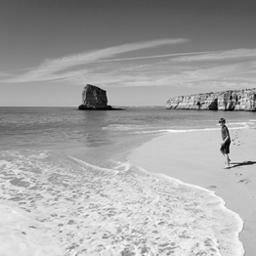}&
        \includegraphics[width=\swidtheight]{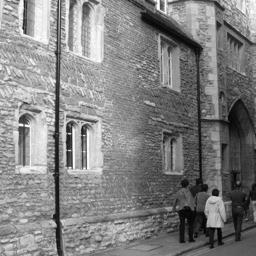}&
        \includegraphics[width=\swidtheight]{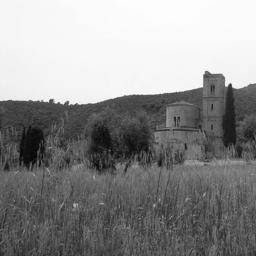}&
        \includegraphics[width=\swidtheight]{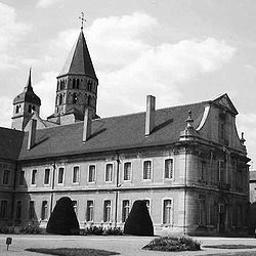}&
				\includegraphics[width=\swidtheight]{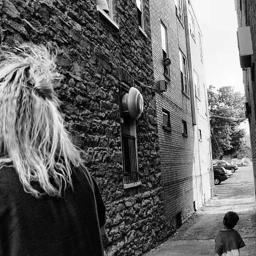}\\
        \rotatebox{90}{Proposed}&
        \includegraphics[width=\swidtheight]{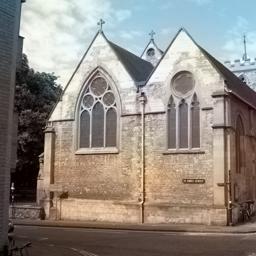}&
        \includegraphics[width=\swidtheight]{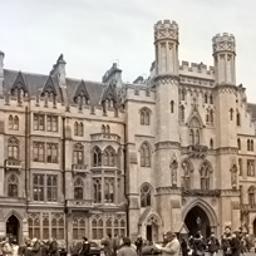}&
        \includegraphics[width=\swidtheight]{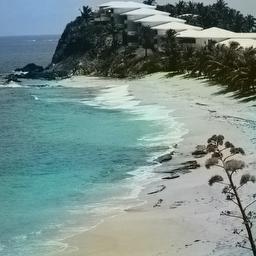}&
        \includegraphics[width=\swidtheight]{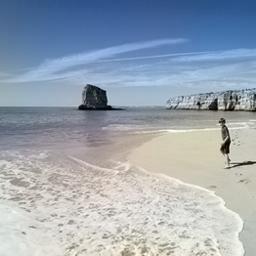}&
        \includegraphics[width=\swidtheight]{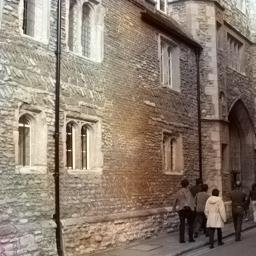}&
        \includegraphics[width=\swidtheight]{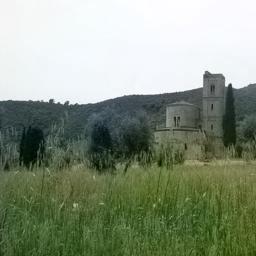}&
        \includegraphics[width=\swidtheight]{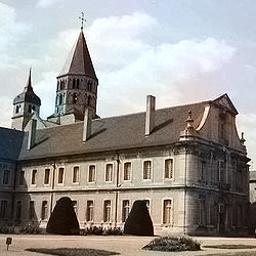}&
				\includegraphics[width=\swidtheight]{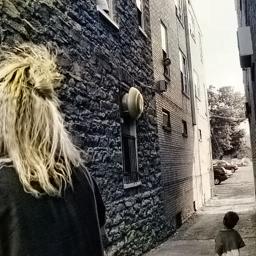}\\
        &25dB&26dB&26dB&24dB&25dB&31dB&28dB&28dB\\
        \rotatebox{90}{Groundtruth}&
		\includegraphics[width=\swidtheight]{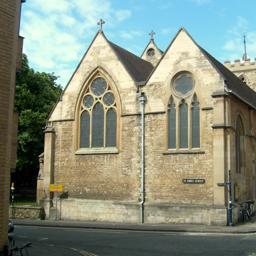}&
        \includegraphics[width=\swidtheight]{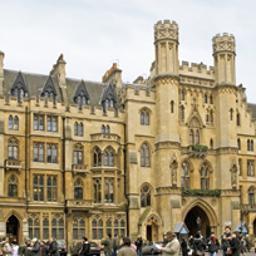}&
        \includegraphics[width=\swidtheight]{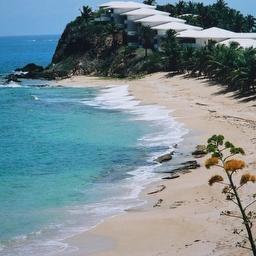}&
        \includegraphics[width=\swidtheight]{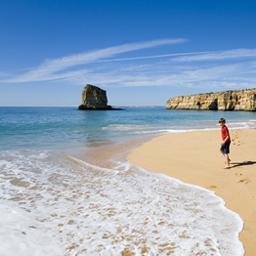}&
        \includegraphics[width=\swidtheight]{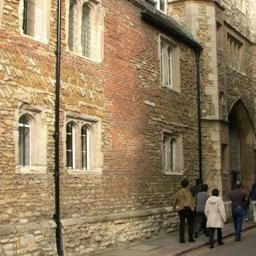}&
        \includegraphics[width=\swidtheight]{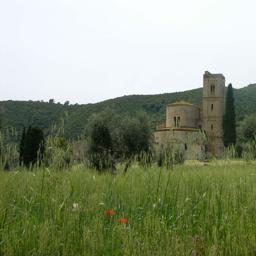}&
        \includegraphics[width=\swidtheight]{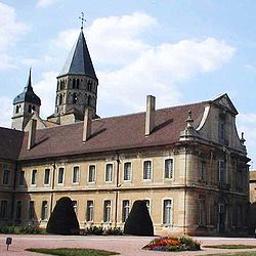}&
        \includegraphics[width=\swidtheight]{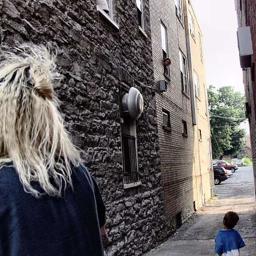}\\
        \end{tabular}
}
    \end{center}
  \caption{Comparison with the ground truth. The 1st/4th/7th rows present the input grayscale images from different categories. Colorization results obtained from the proposed method are presented in the 2nd/5th/8th rows. The 3rd/6th/9th row presents the corresponding ground-truth color images, and the PSNR values computed from the colorization results and the ground truth are presented under the colorized images.}
 \label{Fig:more_results}
\end{figure*}

Figure \ref{Fig:more_results} presents more colorization results obtained from the proposed method with respect to the ground-truth color images \footnote{More colorization results are available at authors' website. \url{http://www.cs.cityu.edu.hk/~qiyang/publications/iccv15/}}. Figure \ref{Fig:more_results} demonstrates that there are almost not visible artifacts in the color images generated using the proposed method, and these images are visually very similar to the ground truth.

\section{Limitations}
\begin{figure}[!ht]
    \begin{center}
        \begin{tabular}{ccc}
        \includegraphics[width=\swidththree]{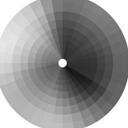}&
        \includegraphics[width=\swidththree]{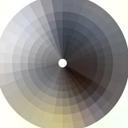}&
        \includegraphics[width=\swidththree]{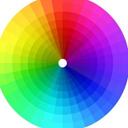}\\
        \includegraphics[width=\swidththree]{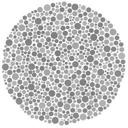}&
        \includegraphics[width=\swidththree]{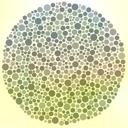}&
        \includegraphics[width=\swidththree]{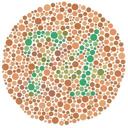}\\
        (a) Input &(b) Our method&(c) Ground truth\\
        \end{tabular}
    \end{center}
  \caption{Limitations of our method. Our method is not suitable for synthetic images and cannot recover the information lost during color to grayscale conversion. Note that the green number in the last row of (c) disappears in the corresponding grayscale image in (a).}
 \label{Fig:limitation}
\end{figure}

The proposed colorization is fully-automatic and thus is normally more robust than the traditional methods. However, it relies on machine learning techniques and has its own limitations. For instance, it is supposed to be trained on a huge reference image database which contains all possible objects. This is impossible in practice. For instance, the current model was trained on real images and thus is invalid for the synthetic image. It is also impossible to recover the color information lost due to color to grayscale transformation. Nevertheless, this is a limitation to all state-of-the-art colorization method. Two failure cases are presented in Figure \ref{Fig:limitation}.

\section{Concluding Remarks}
\label{sec:Conclusion}
This paper presents a novel, fully-automatic colorization method using deep neural networks to minimize user effort and the dependence on the example color images. Informative yet discriminative features including patch feature, DAISY feature and a new semantic feature are extracted and serve as the input to the neural network. An adaptive image clustering technique is proposed to incorporate the global image information. The output chrominance values are further refined using joint bilateral filtering to ensure artifact-free colorization quality. Numerous experiments demonstrate that our method outperforms the state-of-art algorithms both in terms of quality and speed.

\bibliographystyle{IEEEtran}
\bibliography{Reference}

\begin{biography}[{\includegraphics[width=1in,height=1.25in,clip,keepaspectratio]{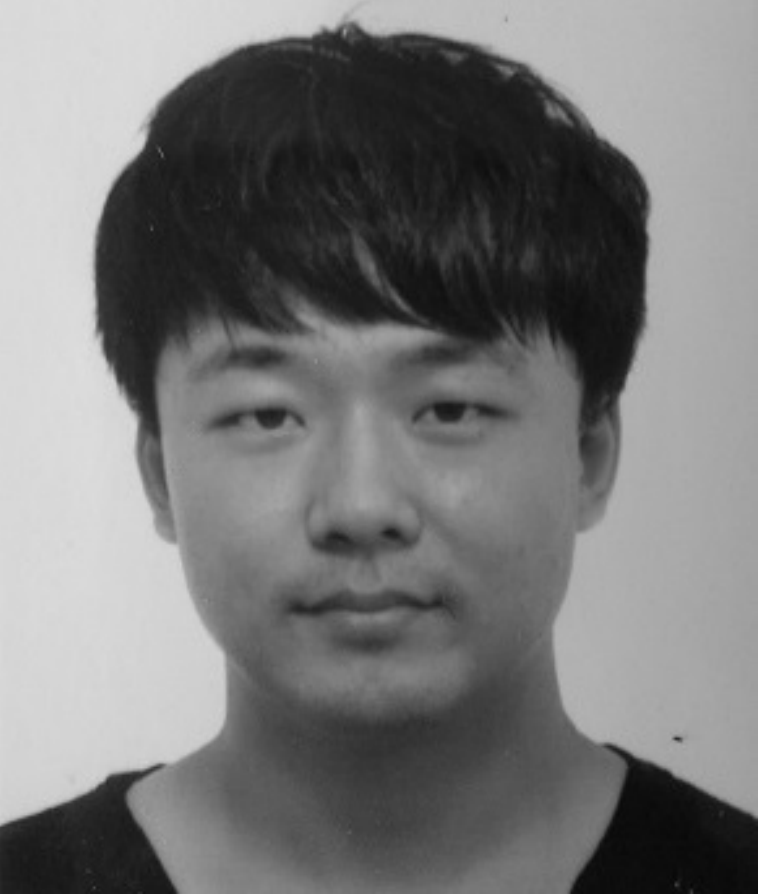}}]{Zezhou Cheng}
received the B.Eng. degree in computer science and technology from Sichuan University, Chengdu, China. He is currently pursuing the Ph.D. degree in the Department of Computer Science and Engineering, Shanghai Jiao Tong University, Shanghai, China. His research interests include image processing, computer vision and machine learning.
\end{biography}
\vspace{-85mm}

\begin{biography}[{\includegraphics[width=1in,height=1.25in,clip,keepaspectratio]{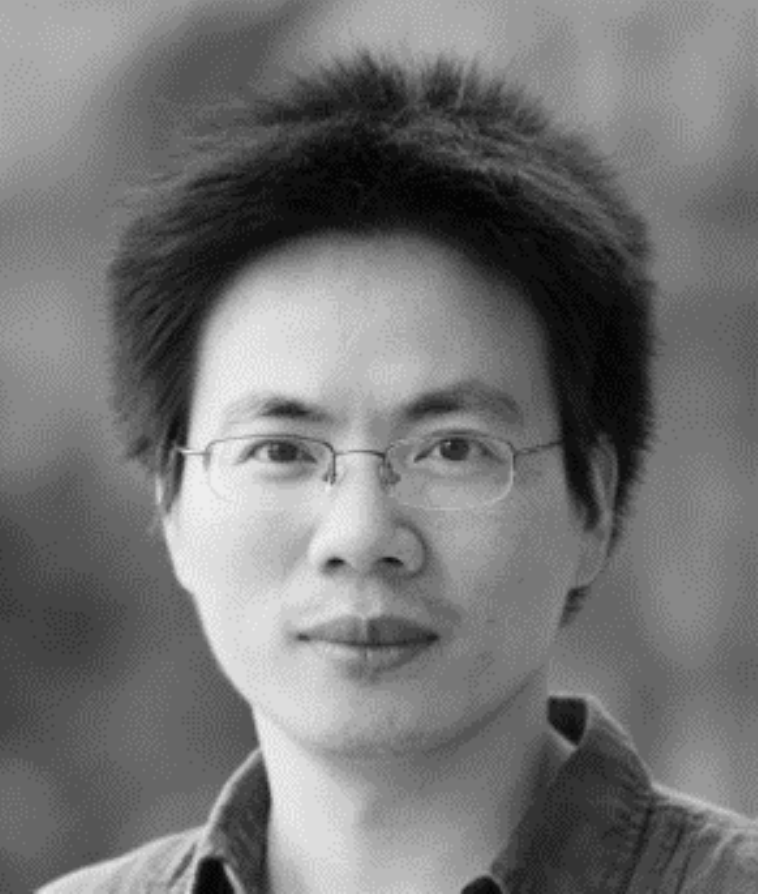}}]{Qingxiong Yang}
is an Assistant Professor in the Department of Computer Science at City University of Hong Kong. He obtained his B.Eng. degree in Electronic Engineering \& Information Science from University of Science \& Technology of China (USTC) in 2004 and PhD degree in Electrical \& Computer Engineering from University of Illinois at Urbana-Champaign in 2010. His research interests reside in Computer Vision and Computer Graphics. He won the best student paper award at MMSP 2010 and best demo at CVPR 2007.
\end{biography}
\vspace{-85mm}

\begin{biography}[{\includegraphics[width=1in,height=1.25in,clip,keepaspectratio]{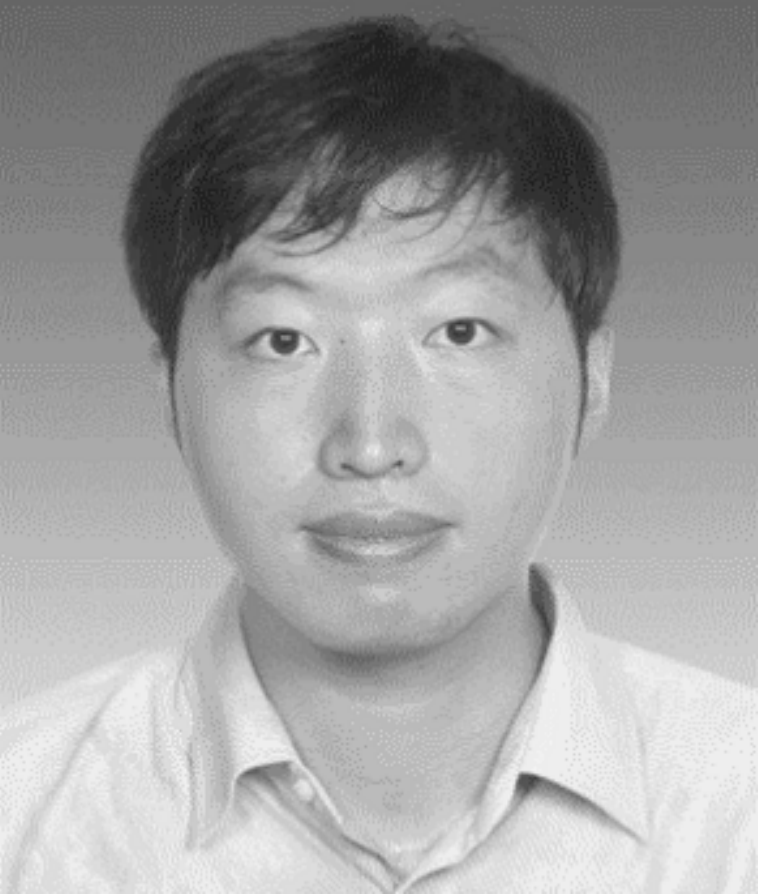}}]{Bin Sheng}
received the B.A. degree in english and
B.E. degree in computer science from Huazhong
University of Science and Technology, Wuhan,
China, in 2004, the M.S. degree in software engineering
from the University of Macau, Macau,
China, in 2007, and the Ph.D. degree in computer
science from The Chinese University of Hong Kong,
Hong Kong, in 2011. He is currently an Associate
Professor with the Department of Computer Science
and Engineering, Shanghai Jiao Tong University,
Shanghai, China. His research interests include
virtual reality, computer graphics, and image-based techniques.
\end{biography}

\end{document}